\documentclass[letterpaper, 10 pt, conference]{ieeeconf}  

\IEEEoverridecommandlockouts                              

\usepackage{graphics} 
\usepackage{graphicx} 
\usepackage{multirow}
\usepackage{caption}

\title{\LARGE \bf
Non-Uniform Spatial Alignment Errors in sUAS Imagery From Wide-Area Disasters
}

\author{Thomas Manzini$^{1*}$, Priyankari Perali$^{1*}$, Raisa Karnik$^{1}$, Mihir Godbole$^{1}$, Hasnat Abdullah$^{1}$, Robin Murphy$^{1}$
\thanks{\textsuperscript{*} Indicates that the authors made equal contributions to this work.}
\thanks{$^{1}$All authors are with the Department of Computer Science and Engineering, Texas A$\&$M University. {Contact: \tt\small{\{tmanzini, perali, raisak, amigo2000, hasnat.md.abdullah, robin.r.murphy\}}@tamu.edu}%
}}

\begin{document}

\maketitle
\thispagestyle{empty}
\pagestyle{empty}

\begin{abstract}

This work presents the first quantitative study of alignment errors between small uncrewed aerial systems (sUAS) georectified imagery and a priori building polygons and finds that alignment errors are non-uniform and irregular, which negatively impacts field robotics systems and human-robot interfaces that rely on geospatial information. 
There are no efforts that have considered the alignment of a priori spatial data with georectified sUAS imagery, possibly because straightforward linear transformations often remedy any misalignment in satellite imagery. 
However, an attempt to develop machine learning models for an sUAS field robotics system for disaster response from nine wide-area disasters 
using the CRASAR-U-DROIDs dataset uncovered serious translational alignment errors.
The analysis considered 21,608 building polygons in 51 orthomosaic images, covering 16787.2 Acres (26.23 square miles), and 7,880 adjustment annotations, averaging 75.36 pixels and an average intersection over union of 0.65.
Further analysis found no uniformity among the angle and distance metrics of the building polygon alignments, presenting an average circular variance of 0.28 and an average distance variance of 0.45 pixels\textsuperscript{2}, making it impossible to use the linear transform used to align satellite imagery. 
The study's primary contribution is alerting field robotics and human-robot interaction (HRI) communities to the problem of spatial alignment and that a new method will be needed to automate and communicate the alignment of spatial data in sUAS georectified imagery.
This paper also contributes a description of the updated CRASAR-U-DROIDs dataset of sUAS imagery, which contains building polygons and human-curated corrections to spatial misalignment for further research in field robotics and HRI.
\end{abstract}

\section{Introduction}
\label{intro}


Small uncrewed aerial systems (sUAS) have been deployed to support disaster response for almost two decades, originally for collecting imagery for reconnaissance and rapid needs assessment missions \cite{murphy2008crew}.
However, starting with Hurricane Harvey \cite{fernandes2018quantitative} and becoming commonplace with Hurricane Ian \cite{manzini2023quantitative}, sUAS have been used to perform wide-area mapping missions, collecting imagery that are combined into a georectified orthomosaic, typically spanning 320 acres and covering entire neighborhoods \cite{manzini2023quantitative, manzini2024crasar}. 
Those sUAS-sourced orthomosaics offer four advantages over satellites for damage assessment: sUAS can be deployed to areas of immediate interest by response personnel without waiting days for satellites to be positioned and oriented; the imagery goes directly to the responders without requiring a high bandwidth internet connection to a central server; the orthomosaics are at a higher resolution which should improve the accuracy of assessments of damage (though covers smaller areas); and sUAS imagery is less expensive to acquire per image. 
Using sUAS imagery instead of satellites to determine the degree of damage to buildings would speed up the strategic understanding of the disaster, allowing decision-makers to deploy immediate aid and accelerate the allocation of state and federal long-term recovery resources. Furthermore, since the sUAS imagery is georeferenced, the damage assessment could even be used to direct tactical assets, perhaps sending sUAS capable of rescue or delivery to a specific location.

\begin{figure}
    \centering
    \includegraphics[width=0.95\columnwidth]{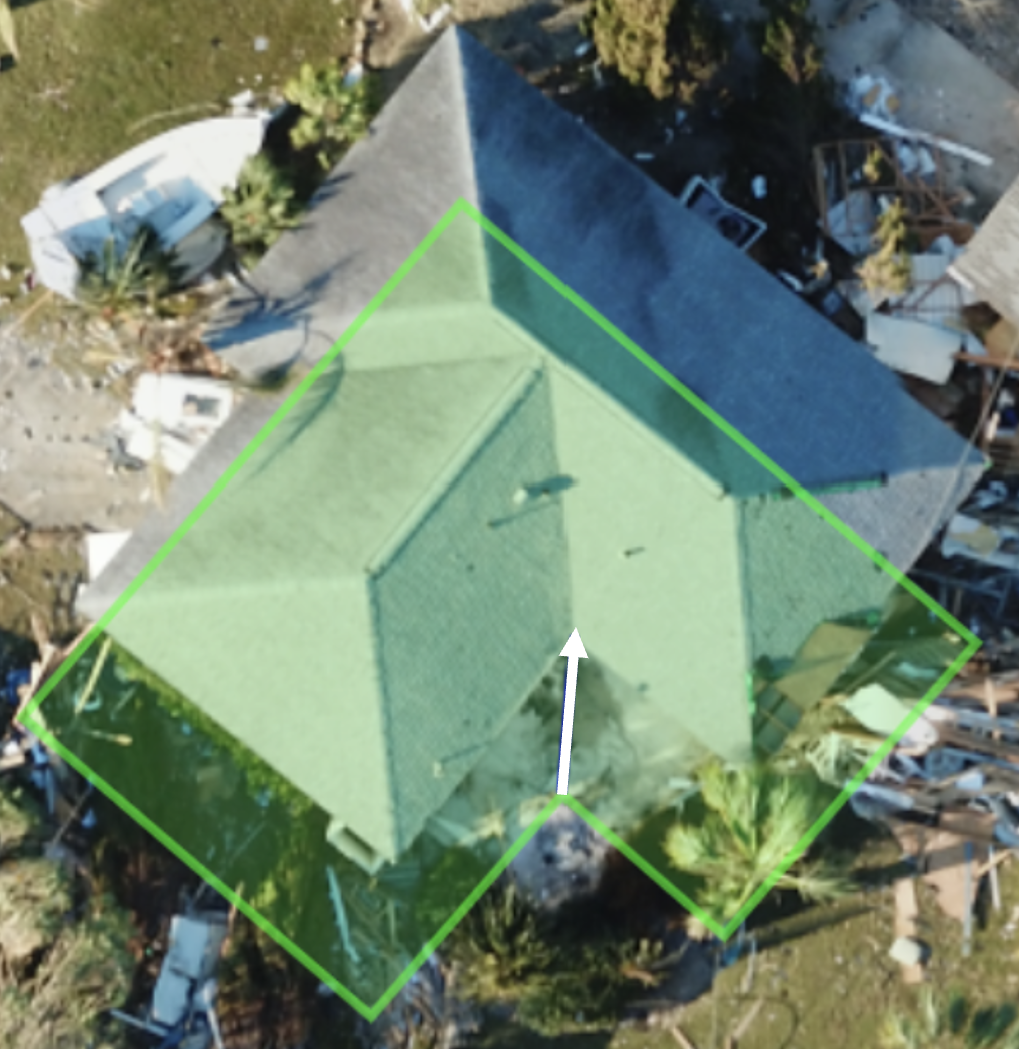}
    \caption{A building footprint polygon (green) out of alignment with the building's structure (image) from Hurricane Ian. The adjustment to align the polygon to the building's structure is shown with a white arrow.}
    \label{fig:adjustment_img}
\end{figure}

sUAS imagery and geospatial data are critical to robotic systems deployed for accelerating disaster response. Notably, path planning and navigation of robotic systems depend on a priori geospatial data \cite{rao2024optimizingstart, rao2024learning, fleischmann2017using, kunz2021localization}. Similarly, machine learning (ML) systems use sUAS imagery and such building polygons to automate building damage assessment \cite{manzini2024crasar}.

Unfortunately, little research has examined how well geospatial data aligns with real-world sUAS-collected imagery. Instead, prior work has been within aligning geospatial data with satellite imagery and has found that satellite imagery is not guaranteed to match the responders' reference polygons because of inconsistencies in the satellite positioning (radiometric distortion) or errors in GPS signals or other sensor systems \cite{congalton2019assessing, Maxar}. This misalignment phenomenon in satellite imagery is often corrected with a single affine transform applied to thousands of buildings \cite{zampieri2018multimodal, vargas2019correcting, gupta2019xbd}.

This paper establishes that spatial misalignment in sUAS orthomosaics is more severe than in satellite imagery, and not as correctable, therefore posing challenges to robotic systems and human-robot interfaces that utilize geospatial data.
By analyzing the CRASAR-U-DROIDs dataset \cite{manzini2024crasar}, this work finds that sUAS imagery does not directly align with the geospatial data used during response operations. Instead, the building polygons are found to be misaligned with sUAS imagery, as shown in Figure \ref{fig:adjustment_img}. 
Further analysis finds no uniformity among the angle and distance metrics of the building polygon alignments as they present a circular variance of 0.44 and an average pixel distance variance of 0.45 pixels\textsuperscript{2}. This lack of uniformity obviates the suitability of a simple affine transform applied to satellite imagery. In response, this work recommends two directions for future work: to automate the correction of non-uniform misalignment and to ensure that misalignment is communicated to decision-makers. These directions are expected to provide a starting point to mitigate the negative implications of the misalignment phenomena if sUAS and sUAS imagery are to be used within real-world operations. 

The core contribution of this work is the quantification of spatial misalignment and the first explicit measurement of the distribution of adjustments necessary to align a priori building polygons with sUAS orthomosaic imagery. 
The remaining paper is organized as follows. 
Section \ref{sec:related_work} sets the foundation for the quantitative analysis by describing the process of generating georeferenced sUAS orthomosaics and reviewing the existing literature on obtaining photogrammetric accuracy and on ML incorporating geospatial data. 
Section \ref{sec:approach} provides details of the CRASAR-U-DROIDs dataset and the alignment process, explaining the process of constructing sUAS orthmosaics from nine disasters, aligning damage with 21,608 building polygons, and how manual adjustments align the polygons with the imagery. The manual adjustments are key elements of the quantitative analysis presented in Section \ref{sec:results}. A discussion of the limitations of the analysis and its implications follows in Section \ref{sec:discussion}, concluding that spatial alignment is a major barrier and providing direction for future research in using sUAS imagery for robotic systems and human-robot interfaces (Section \ref{sec:conclusions}). 

\section{Related Work}
\label{sec:related_work}

To provide the necessary background and context for understanding the consequences of spatial misalignment, this section discusses the potential sources of misalignment. It then reviews relevant research showing the alignment problem in sUAS orthomosaic imagery is similar to research in improving photogrammetry accuracy; however, those methods assume ground control points, which are not always available during disasters. This section also reviews research in training ML models from satellite imagery. Those efforts
have already identified that coincident satellite imagery will have spatial errors that need to be managed when training ML models, but none of these have explicitly measured the distribution of alignment errors, leading this paper to craft a novel analysis method presented in Section~\ref{sec:approach}.

Before discussing the literature, it is helpful to establish the terminology that will be used throughout the remainder of the paper.
Different communities may refer to this work as ``registration" \cite{garcia2020pix2streams}, ``alignment" \cite{vargas2019correcting, manzini2023quantitative, zampieri2018multimodal},
``image shifting" \cite{gupta2019xbd}, a study of ``reconstruction accuracy," ``spatial accuracy," or ``positional precision" \cite{barbasiewicz2018analysis, dare2002digital, liba2015accuracy, ludwig2020quality, azim2019manual, hung2019positional}.
Due to the sparse and non-visual nature of building polygon data and the way it is being utilized in this work, ``alignment" represents the most appropriate term.

\begin{figure*}
    \centering
    \includegraphics[width=0.95\textwidth]{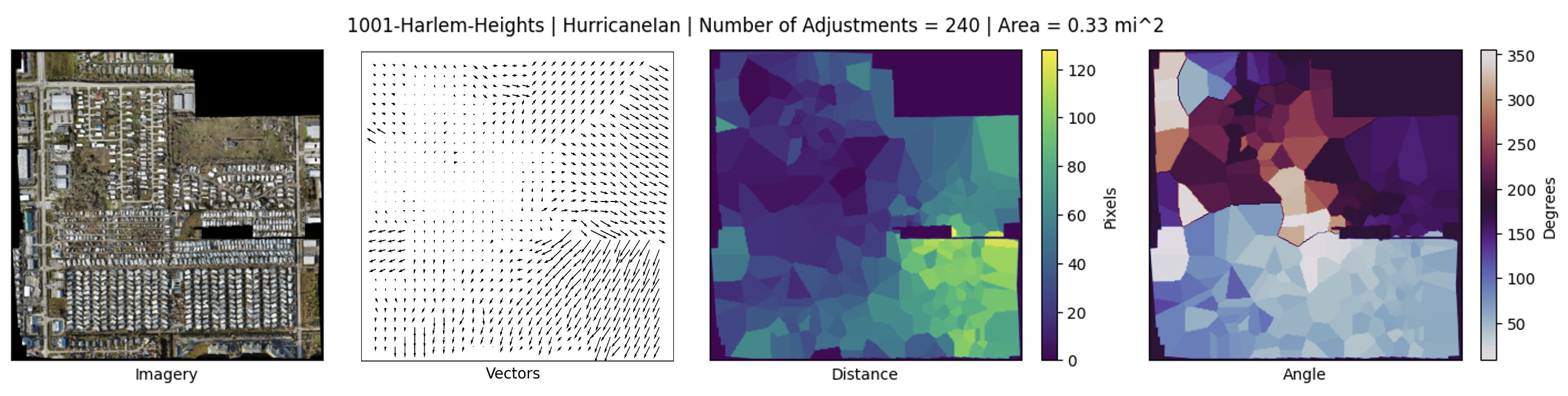}
    \vspace{-0.2cm}
    \caption{Visualization of the adjustments for the 1001-Harlem-Heights orthomosaic from Hurricane Ian. Plotted together is the raw imagery [left], the vector field [middle-left], with a colored visualization for distance [middle-right] and angles [right]. Note the discontinuous and non-uniform distribution of adjustments.}
    \label{fig:heatmap}
\end{figure*}

\subsection{Sources of alignment error}
\label{subsec:source_alignment_errors}

Alignment errors between sUAS imagery and spatial data considered here are derived from five sources: satellite imagery acquisition, polygon generation, scale variance between sUAS and satellite, sUAS GPS noise, and errors induced by the orthomosaic generation process. Each is discussed separately below. 


Typically, building polygons are generated from satellite imagery \cite{van2018spacenet, gupta2019xbd}. The creation of these polygons presents three sources of error. First, the imagery from which the building polygons are derived may be misaligned itself. Second, the process by which the building polygons are derived may introduce further noise depending on the specifics of the system. Third, satellites and sUAS do not capture imagery at the same ground sampling distances (GSD). Satellites capture imagery at a wide variety of GSDs; however, in disaster response, it is common to use GSDs in the 1 m/px to 30 cm/px range \cite{MaxarSpatialResolution, gupta2019xbd}. While sUAS can generate far higher resolutions, with the imagery in this work averaging 3.74 cm/px \cite{manzini2024crasar}.
Compared to higher-resolution sUAS imagery, satellite imagery is intrinsically less precise, creating the potential for discrepancies between the two \cite{manzini2025now}.

The spatial coordinate system of sUAS orthomosaic imagery is based in part on the sUAS's GPS, which is subject to error that arises from atmospheric conditions \cite{williams1998integrated} and interference with urban structures \cite{merry2019smartphone}. When there is GPS error, the overlaid building polygons will be based on positional inaccuracies, creating misalignment.  

The orthomosaic generation process is another area where errors may be introduced in sUAS imagery. After sUAS imagery is collected, it is used to construct an orthomosaic, shown in Figure \ref{fig:heatmap} on the far left. During construction, the orthomosaic generation software may manipulate each image by rotating, scaling, skewing, and translating. This process represents another opportunity for errors to arise \cite{dare2002digital}.

\subsection{Related Work in Photogrammetry Accuracy}

While there has been substantial work in developing and measuring the accuracy of data products generated by photogrammetry software \cite{barbasiewicz2018analysis, dare2002digital} and further work measuring the accuracy when imagery is captured by sUAS \cite{liba2015accuracy, ludwig2020quality, azim2019manual, hung2019positional}, this paper is interested in the degree to which the orthomosaics align with programmatically generated building polygons obtained from satellite imagery.
The common approach in these papers is to use ground control points (GCPs)-points on the earth whose position is known precisely to measure the accuracy of, and in some cases further refine, the generated data product or orthomosaic \cite{liba2015accuracy, ludwig2020quality, azim2019manual, hung2019positional}. 
While the content of this paper can be seen as a neighbor to this area of work, it is different as this work makes no use of GCPs and is not interested in the underlying accuracy of orthomosaics, instead measuring the differences between orthomosaic imagery and spatial data.

\subsection{Related Work in Polygons and Geospatial ML}

Prior works \cite{gupta2019xbd, maiti2022effect, vargas2019correcting} have already identified that coincident satellite imagery will have spatial errors that need to be managed when training ML models; however, none of these have explicitly measured the distribution of alignment errors, which is the focus of this paper. Two studies have measured the impact of alignment quantitatively \cite{vargas2019correcting, maiti2022effect}. However, they only measured the impact on ML model performance rather than the misalignment phenomenon itself.

Four strategies have been considered for the management of alignment errors. First, and most simply, in satellite imagery, a uniform translation to all building polygons has been considered \cite{gupta2019xbd}. 
A second strategy is, in satellite imagery, to use, \cite{vargas2019correcting, zampieri2018multimodal, girard2019aligning}, more complex techniques, CNN models, and Markov Random Fields. However, it is unclear how effective these techniques will be in sUAS imagery as they either enforce affine transformation constraints during optimization \cite{zampieri2018multimodal, girard2019aligning} or rely on computing a graph structure to represent nearby buildings, so they can be translated together uniformly \cite{vargas2019correcting}.
A third strategy was to use ``off-the-shelf image registration algorithms" to automatically align spatial polygons with imagery \cite{garcia2020pix2streams}, though it is unclear how these algorithms were applied. 
Finally, a fourth effort subverted this problem by maintaining an internal set of polygons that they constructed and then queried using a nearest neighbor search with a 20-meter threshold \cite{robinson2023rapid} of a target polygon.

\section{Approach}
\label{sec:approach}

The approach taken in this work is to analyze the adjustments manually applied to building polygons necessary to align them with the orthomosaic imagery \cite{manzini2024crasar}. The adjustment angles and distances are aggregated, and misalignment is quantified using the statistical measures, mean and variance, 
and intersection over union (IoU) of the aligned and unaligned building polygons. This section details the orthomosaic imagery, building polygons, and adjustment generation process, with analysis in Section \ref{sec:results}. 

\subsection{sUAS Orthomosaic Imagery}

The imagery used within this analysis to align the building polygons consisted of 51 orthomosaics collected from sUAS at 9 wide-area disasters in the United States, sourced from the CRASAR-U-DRIODs dataset \cite{manzini2024crasar}. Orthomosaic imagery (shown in Figure \ref{fig:heatmap} on the far left) is a collection of images taken from an aerial camera that have been stitched together to produce an image that is pixel-aligned with longitude and latitude. This imagery includes six hurricanes (Hurricane Ian, Hurricane Harvey, Hurricane Michael, Hurricane Ida, Hurricane Idalia, and Hurricane Laura), the Mayfield Tornado, the Musset Bayou Fire, and the Kilauea Eruption.  

The imagery considered was collected by 9 distinct models of sUAS. In descending order of prevalence, these were the DJI Mavic 2 (16 orthomosaics), SenseFly eBee X (12 orthomosaics), DJI Matrice 30 (6 orthomosaics), DJI Mavic Pro (5 orthomosaics), DJI Phantom 4 (4 orthomosaics), DJI Matrice 600 (3 orthomosaics), DJI Matrice 300 (3 orthomosaics), Wingtra WingtraOne Gen II (2 orthomosaics), Parrot Anafi (1 orthomosaic). 
Orthomosaics were generated using Pix4D React (98\%), and Agisoft Metashape (2\%).

\subsection{Manual Adjustment Method}
\label{sec:adj_method}

The manual adjustment method consisted of annotating the imagery with adjustments and using the annotated adjustments to align the building polygons provided by the Microsoft Building Footprints Dataset \cite{MicrosoftBuildingFootprints}.
An \emph{adjustment} refers to a line that is drawn on an image by an annotator. Each adjustment is a line drawn from a building polygon vertex to the true location of the vertex in the image; an example of this is shown in Figure \ref{fig:adjustment_img}. Whereas an \emph{alignment} refers to the translation done to the building polygon itself. During the annotation process, adjustments were provided for 36\% of all building polygons. The remaining, unannotated, building polygons were aligned based on the following logic.

When an alignment is needed in a given area, the nearest adjustment is taken and applied.
This process forms a vector field as shown in Figure \ref{fig:heatmap}.
More formally, this vector field is generated by a piecewise linear combination of all the adjustments, where the nearest adjustment is selected to populate the vector field. 
All vector fields were manually reviewed for correctness by the authors. This was done by inspecting the alignment of all buildings in all orthomosaics and manually adding or correcting adjustments until all buildings were correctly aligned. The adjustment collection and curation process is described further in \cite{manzini2024crasar}.

\begin{figure}
    \centering
    \includegraphics[width=0.9\columnwidth]{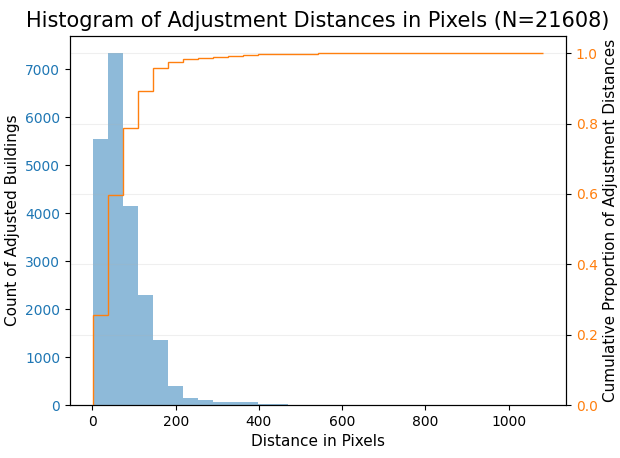}
    
    \vspace{0.0cm} 
    
    \centering
    \begin{tabular}{|l|llllll|}
    \hline
    Percentile    & \multicolumn{1}{l|}{p10} & \multicolumn{1}{l|}{p25} & \multicolumn{1}{l|}{p50} & \multicolumn{1}{l|}{p75} & \multicolumn{1}{l|}{p90} & p99   \\ \hline
    Dist$_{px}$ & \begin{small}17.9 \end{small}                    & \begin{small}35.0 \end{small}                    & \begin{small}59.9 \end{small}                    & \begin{small}99.4 \end{small}                    & \begin{small}147.3\end{small}                    & \begin{small}324.0\end{small} \\ \hline
    Dist$_{cm}$ & \begin{small}67.2 \end{small}                    & \begin{small}127.5\end{small}                    & \begin{small}200.7\end{small}                    & \begin{small}318.3\end{small}                    & \begin{small}481.8\end{small}                    & \begin{small}843.4\end{small} \\ \hline
    \end{tabular}
    
    \caption{Histogram and CDF of alignment distances in pixels. Percentiles are shown below for pixels and centimeters.}
    \label{fig:distance-cdf}
\end{figure}

\section{Analysis of Misalignment}
\label{sec:results}
As mentioned in Section \ref{sec:approach}, the building polygon alignments' angles and distances were measured and aggregated, and their misalignments were quantified through statistical, 
and IoU measures. 
The distances of the alignments were computed for both the pixel and centimeter distances of the adjustments.  
This section presents the analysis of the results with a discussion following in Section \ref{sec:discussion}. 

\subsection{Mean}
The mean distance and angle of aligned building polygons were used to estimate the average case for alignment errors in orthomosaic imagery. 
The analysis finds that, across all 51 orthomosaics, the mean alignment angle was 277.98 degrees, and the mean alignment distance was 75.36 px or 246.11 cm. The mean alignment angle for each orthomosaic ranged from 10.97 degrees to 356.12 degrees. The mean alignment distance for each orthomosaic ranged from 14.05 px to 240.97 px or 71.9 cm to 839.7 cm. The distance component was further analyzed by measuring the distribution of all alignments together, as shown in Figure \ref{fig:distance-cdf} with percentiles added for reference. The distribution has a median distance of 59.5 px and 200.7 cm but features a long tail of alignments.  


\begin{figure}
    \centering
    \includegraphics[width=0.9\columnwidth]{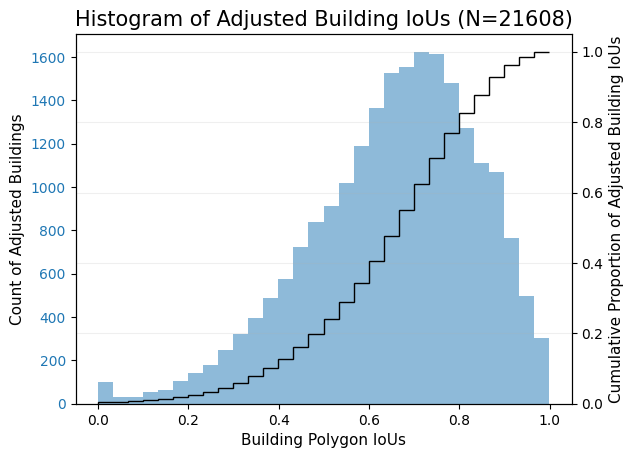}
    
    \vspace{0.0cm} 
    
    \centering
\begin{tabular}{|l|llllll|}
\hline
Percentile & \multicolumn{1}{l|}{p10} & \multicolumn{1}{l|}{p25} & \multicolumn{1}{l|}{p50} & \multicolumn{1}{l|}{p75} & \multicolumn{1}{l|}{p90} & p99  \\ \hline
IoU        & \begin{small}0.40\end{small}                     & \begin{small}0.54\end{small}                     & \begin{small}0.68\end{small}                     & \begin{small}0.79\end{small}                     & \begin{small}0.88\end{small}                     & \begin{small}0.97\end{small} \\ \hline
\end{tabular}
    
    \caption{Histogram and CDF of the Intersection over Unions (IoUs) between unaligned and aligned building polygons.}
    \label{fig:iou_cdf}
\end{figure}

\subsection{Variance}
The variances of distances and angles of aligned building polygons were used to characterize the width of the distribution of alignment errors in orthomosaic imagery.
The average circular variance across all orthomosaics is 0.28, and the average distance variance is 0.45 px\textsuperscript{2}. The circular variance ranged from 0 to 0.83. The average distance variance ranged from 0 px\textsuperscript{2} to 0.95 px\textsuperscript{2}. 
It is worth noting that the minimum average circular variance of 0 and average distance variance of 0 px\textsuperscript{2} are associated with the 090402-DMS-Assessment-Sienna-Village.geo.tif orthomosaic that has a limited number of buildings (n=16) and adjustment annotations (n=2). This was the only orthomosaic with such a variance.


\subsection{Intersection over Union}

The intersection over the union (IoU) of the aligned and unaligned building polygons was explored to contextualize the scale of adjustments with the scale of the building polygons.
The IoUs for all building polygons were taken and aggregated per orthomosaic. The average IoUs ranged from 0.24 to 0.87, averaging 0.65. 

The median IoU across all building polygons considered in this work is 0.68, indicating that the misalignment phenomenon results in building polygons that are apart from the locations of the buildings in the imagery. The distribution of all building polygon IoUs is shown in Figure \ref{fig:iou_cdf}.

\section{Discussion}
\label{sec:discussion}

The analysis results presented in Section \ref{sec:results} are indicative of varying distributions, suggesting a need for an in-depth interpretation of them, and pushing for future work to address the needs of alignment.
This section further discusses the analysis results and implications for future work. 

\subsection{Limitations}
Due to the limited availability of imagery and spatial data, this work is hindered by four limitations. First, this analysis is limited to alignment translation errors, meaning measurements of skew, rotational, or polygon shape errors were not included. Though these additional misalignment errors were rarely encountered, anecdotally, polygon shape errors represent the largest proportion of unaddressed misaligned building polygons, followed by rotational errors.
Second, this work analyzes alignment errors within four types of disasters (hurricanes, wildfires, volcanic eruptions, and tornadoes), which do not capture alignment errors within other types of disasters (e.g., earthquakes, tsunamis). There is the potential that alignment errors will vary among different disaster types.
Third, this work is limited to orthomosaics that were generated by Pix4D React and Agisoft Metashape. The misalignment phenomenon may depend in part on the georectification techniques implemented by mapping software. 
Lastly, this analysis only considers alignment errors within a priori building polygons. Other spatial data may reveal differing misalignments, but it is expected that this work will be a starting point for analysis of other spatial data.

\subsection{Interpretation of Results}
Considering the results presented in Section \ref{sec:results}, the evident trend is a lack of uniformity in the types of errors that are encountered. 
It is observed that these adjustments are non-uniform within orthomosaics as observed in Figure \ref{fig:heatmap}. Moving to Figure \ref{fig:heatmap}, these errors are spatially non-linear and discontinuous in both adjustment distance and angle. As mentioned in Section \ref{sec:adj_method}, the adjustments are obtained via a linear piecewise combination by selecting the nearest adjustment to a given building.

The measurements of average circular variance are the most relevant to this discussion. Efforts to align spatial data within satellite imagery with spatial data assume an affine transformation \cite{zampieri2018multimodal, gupta2019xbd, maiti2022effect, vargas2019correcting}. However, the average circular variance across orthomosaics collected ranges from 0 to 0.83.
These values suggest a single linear transformation of the locations of building polygons will not be sufficient to capture the variety of alignment errors observed.

The measurements of IoU are also particularly relevant, as these measures show that the misalignment phenomena is more than minor noise. Instead, misalignment accounts for a median of 0.68 IoU of buildings being covered between aligned and unaligned building polygons. Such a variation in IoU would account for a substantial performance change in ML models as evidenced in \cite{maiti2022effect, vargas2019correcting}.

\subsection{Implications for Field Robotics and HRI}
Field robotic systems and HRI that depend on the combination of spatial data and sUAS imagery are negatively impacted by the non-uniform misalignment found within this analysis, posing two fundamental implications when deploying field robotic systems within real-world disasters. Without reliable methods to correct the non-uniform misalignment between spatial data and sUAS imagery, real-world deployments of field robotics systems and data products for decision-makers will impede operational benefit.

Misalignment poses a risk of performance degradation within field robotics systems. As mentioned earlier, field robotic systems utilize a combination of spatial data and sUAS imagery for path planning \cite{rao2024optimizingstart, rao2024learning} and navigation \cite{fleischmann2017using, kunz2021localization}. Without correct alignment, the a priori spatial information (e.g., building locations) provided to path planning and navigation algorithms risks containing inaccuracies, resulting in performance degradations. For instance, consider a path planning algorithm using the ergodic metric, optimizing spatial coverage between high and low information regions by aerial and ground robots \cite{rao2024optimizingstart, rao2024learning}. Such an algorithm would be hindered by misalignment between a priori information (e.g., building locations) because the geospatial data does not accurately describe high-information regions. 

Misalignment poses a risk of miscommunication to decision-makers through HRI interfaces. Decision makers who use field robotic and ML systems that utilize a combination of spatial data and sUAS imagery risk making incorrect decisions when misalignment has not been corrected. For one, operators who task field robots may rely on spatial data or in-situ imagery of the scene to task and direct robots, as observed in \cite{manzini2023quantitative}. As a result, misalignment may contribute to incorrect allocation of robots or confusion in the field \cite{honig2018understanding}. 

To further complicate this, field robotic systems and HRI interfaces may rely on the ML model outputs that also utilize the same misaligned spatial data and sUAS imagery. As discussed within section \ref{sec:related_work}, building performant ML systems that leverage spatial data depends upon the accurate alignment of that spatial data with the imagery for effective training and inference. For example, a 2-meter translation error during ML training on crewed aircraft imagery can induce as much as a 10\% accuracy loss during inference \cite{maiti2022effect}. Translation noise during inference on satellite imagery has been shown to decrease performance by at least 0.546 F-score (an 83\% relative decrease in F-score) \cite{vargas2019correcting}. These dynamics are dependent on ML model architecture and objective, but model performance consistently improves with reducing alignment errors \cite{maiti2022effect, vargas2019correcting}. However, due to insufficient approaches \cite{vargas2019correcting, zampieri2018multimodal, girard2019aligning} to correct the non-uniform misalignment for ML systems, they pose the risk of increased incorrect predictions (e.g., increased incorrect damage label predictions), which in turn increase the risk of degraded performance for path planning and navigation that consume these predictions as a priori information and risk miscommunication to decision-makers. 

\section{Conclusion}
\label{sec:conclusions}
This work presented the first quantitative study, explicitly measuring the distributions of building polygon alignment, and contributed a description of the spatial data within the CRASAR-U-DRIODs dataset. Based on the 51 orthomosaics, spanning nine wide-area disasters and consisting of 21,608 building polygons, involved in this analysis, the study found that, with a circular between 0 to 0.83 and an average distance alignment variance of 0.45 pixels\textsuperscript{2}, the alignment distributions are non-uniform presenting a barrier to field robotic systems and HRI interfaces. Given the impacts of these misalignments on field robotics and HRI,
future work is needed to mitigate the implications of misalignment if sUAS systems are to provide operational benefit. The recommendations for future work are as follows. 
\begin{itemize}
    \item Field robotics and ML researchers should work to automate the reconciliation of misalignments for sUAS imagery to be better interpreted in real-time \cite{zampieri2018multimodal, girard2019aligning}. 
    \item HRI researchers should work to communicate the misalignment phenomenon with the interfaces presented to decision-makers.
\end{itemize}
As sUAS systems and their imagery become more integrated in disaster response, the field robotics and HRI community should be aware of misalignment and its implications.






\section*{ACKNOWLEDGMENT}
This material is based upon work supported by the AI Research Institutes Program funded by the National Science Foundation under the AI Institute for Societal Decision Making (NSF AI-SDM), Award No. 2229881, and under ``Datasets for Uncrewed Aerial System (UAS) and Remote Responder Performance from Hurricane Ian" Award No. 2306453. Acknowledgment is given to CRASAR and David Merrick for imagery acquisition and supporting information.


\bibliographystyle{IEEEtranS}
\bibliography{references}

\end{document}